# MEKF Ignoring Initial Conditions for Attitude Estimation Using Vector Observations

Lubin Chang

***Abstract*—In this paper, the well-known multiplicative extended Kalman filter (MEKF) is re-investigated for attitude estimation using vector observations. From the Lie group theory, it is shown that the attitude estimation model is group affine and its error state model should be trajectory-independent. Moreover, with such trajectory-independent error state model, the linear Kalman filter is still effective for large initialization errors. However, the measurement model of the traditional MEKF is dependent on the attitude prediction, which is therefore trajectory-dependent. This is also the main reason why the performance of traditional MEKF is degraded for large initialization errors. Through substitution of the attitude prediction related term with the vector observation in body frame, a trajectory-independent measurement model is derived for MEKF. Meanwhile, the MEKFs with reference attitude error definition and with global state formulating on special Euclidean group have also been studied, with main focus on derivation of the trajectory-independent measurement models. Extensive Monte Carlo simulations and field test of attitude estimation implementations demonstrate that the performance of MEKFs can be much improved with trajectory-independent measurement models.**

## I. Introduction

EXTENDED Kalman filter (EKF) has been widely used for attitude estimation [1-3]. With quaternion as the attitude parameterization, the resultant EKF is known as multiplicative extended Kalman filter (MEKF) [4, 5]. In MEKF, the quaternion is used for the global attitude propagation and a local unconstrained three-dimensional parameterization is used for local attitude error estimation in filtering recursion. The MEKF has been widely approved as the workhorse of real-time attitude estimation using vector observations [6]. However, the MEKF may become very difficult to tune and even prone to divergence when it has not been well initialized. In order to relax restrictions on valid initialization region, some advanced filtering methods have been proposed [7, 8]. These advanced filtering methods can indeed exceed MEKF for large initialization errors, but at the cost of increasing computational burden. In [9], a new frame-consistent EKF termed as geometric EKF (GEKF) is proposed for spacecraft attitude estimation. In GEKF, an attitude error related gyroscope bias error is defined based on common coordinates perspective. The GEKF can outperform MEKF for large initialization errors with similar computational cost. In recent few years, the invariant extended Kalman filter (IEKF) rooted in the matrix Lie group theory is investigated for attitude or pose estimation in inertial based applications [10-15]. In [11], it is argued that IEKF exhibits superior convergence and robustness properties over MEKF, due to that its error state model does not depend on the state estimate. In [16], two quaternion IEKF are investigated for spacecraft attitude estimation based on the invariance theory. It is pointed out that the MEKF can be viewed as a minor variant of the quaternion left IEKF (QLIEKF). In [17], it is pointed out that both the GEKF in [6] and quaternion right IEKF (QRIEKF) in [16] can be derived through formulating the attitude and gyroscope drift bias together as element of special Euclidean group $\mathbb{SE}(3)$.

It is generally recognized that the performance degradation of MEKF with large initial attitude error owes much to the local linearization. In other words, the local linearized error state model is not applicable for case with large attitude error. However, such conclusion may not hold true universally. From the Lie group theory, if the group state dynamical model is group affine, the corresponding linear vector error state model will be trajectory independent. Moreover, the nonlinear group error can be recovered exactly from the linear vector error [12, 18]. In other words, the linear approximation for group affine will not lose accuracy. In this paper, it is demonstrated that the attitude model formulating on the special orthogonal group $\mathbb{SO}(3)$ is group affine and therefore, the MEKF possess the potential to handle large initialization errors. Based on the aforementioned discussion, with a small modification of the measurement transition matrix of the existing MEKF, its performance can be much improved. Moreover, the MEKF with reference attitude error and MEKF with global state formulating on $\mathbb{SE}(3)$ have also been revisited and investigated from the perspective of group affine.

The remainder of this paper is organized as follows: Section II presents the dynamic model for attitude estimation using vector observations. The group affine property of the model has also been analyzed, which provides the theoretical basis for the probability of handling large initialization errors with the linear state-space models. In Section III, the state-space models with both body frame and reference frame attitude errors have been presented. A modified measurement model which is trajectory-independent is proposed for body frame attitude error definition.

The paper was supported in part by National Natural Science Foundation of China (61873275).

The author with the College of Electrical Engineering, Naval University of Engineering. (e-mail: changlubin@163.com).

Meanwhile, through transformation of the innovation, the measurement model can also be transformed into trajectory-independent for reference frame attitude error definition. In Section IV, through formulating the attitude and gyroscope drift bias as compact element of $\mathbb{SE}(3)$ , the corresponding vector error state-space models are presented. In Section V, the explicit filtering procedures have been presented with different state-space models. The focus of this section is put on the global state update procedures corresponding to their respective error definitions. Numerical simulations are given in Section VI to evaluate the performance of these different state-space models involved in the MEKF and field test is carried out in Section VII to further evaluate the superiority of the trajectory-independent state-space model. Finally, conclusions are drawn in Section VIII.

## II. Group Affine Analysis of Attitude Estimation Model

### A. System Model

The attitude kinematics model in terms attitude matrix is given by [19]

$$\dot{\mathbf{A}}(t) = -\left[\boldsymbol{\omega}(t)\times\right]\mathbf{A}(t) \tag{1}$$

where $\boldsymbol{\omega}$ is the angular rate measured by the gyroscopes and governed by

$$\begin{aligned}\tilde{\boldsymbol{\omega}} &= \boldsymbol{\omega} + \boldsymbol{\beta} + \boldsymbol{\eta}_{\mathbf{v}}\\ \dot{\boldsymbol{\beta}} &= \boldsymbol{\eta}_{\mathbf{u}}\end{aligned} \tag{2}$$

where $\boldsymbol{\beta}$ is the constant gyroscope bias. $\boldsymbol{\eta}_{\mathbf{v}}$ and $\boldsymbol{\eta}_{\mathbf{u}}$ are assumed to be zero-mean, Gaussian white-noise processes.

In (1), the attitude matrix $\mathbf{A}(t)$ denotes the attitude transformation from inertial frame to body frame and the angular rate $\boldsymbol{\omega}(t)$ can be measured direct by the gyroscopes. However, for attitude estimation with local-level frame as the reference frame on the earth surface, $\boldsymbol{\omega}(t)$ contains earth rotation and movement of the local-level frame over Earth's curvature. In such case, the earth rotation and movement of the local-level frame over Earth's curvature can be handled through ingenious attitude decomposition if the body position and velocity are known [20].

The measurement model corresponding to the vector observation is given by

$$\mathbf{b}_k = \mathbf{A}(t_k)\mathbf{r}_k + \boldsymbol{\upsilon}_k \tag{3}$$

where $\mathbf{r}_k$ is the reference vector observation at time instant $t_k$ and $\mathbf{b}_k$ is the observation vector with measurement noise $\boldsymbol{\upsilon}_k$ . Here only one vector observation is considered and the generalization to case with multiple vector observations is straightforward.

### B. Group Affine Analysis of Spacecraft Attitude Estimation Model

Given a group state model

$$\dot{\chi} = f_{\mathbf{u}}(\chi) \tag{4}$$

where $\chi \in \mathcal{G}$ is the state on group $\mathcal{G}$ . $\mathbf{u}$ are the model inputs. From the Lie group theory, if the group state model (4) satisfies the following equation, it is group affine [9].

$$f_{\mathbf{u}}(\chi_1\chi_2) = f_{\mathbf{u}}(\chi_1)\chi_2 + \chi_1 f_{\mathbf{u}}(\chi_2) - \chi_1 f_{\mathbf{u}}(\mathbf{I}_{nd})\chi_2 \tag{5}$$

where $\chi_1, \chi_2 \in \mathcal{G}$ .

It is known that attitude matrix $\mathbf{A}(t)$ is on group $\mathbb{SO}(3)$ . Next, we will demonstrate that the attitude kinematics (1) is group affine. In the following demonstration, the measurement errors of gyroscopes are not considered and the involved variables' time dependences on are omitted for brevity temporarily.

For model (1), the group state is $\mathbf{A}$ and the input is $\boldsymbol{\omega}$ . According to (1), we can get

$$f_{\mathbf{u}}(\mathbf{A}_1\mathbf{A}_2) = -(\boldsymbol{\omega}\times)\mathbf{A}_1\mathbf{A}_2 \tag{6a}$$

$$f_{\mathbf{u}}(\mathbf{A}_1)\mathbf{A}_2 = -(\boldsymbol{\omega}\times)\mathbf{A}_1\mathbf{A}_2 \tag{6b}$$

$$\mathbf{A}_1 f_{\mathbf{u}}(\mathbf{A}_2) = -\mathbf{A}_1(\boldsymbol{\omega}\times)\mathbf{A}_2 \tag{6c}$$

$$\mathbf{A}_1 f_{\mathbf{u}}(\mathbf{I}_{3\times3})\mathbf{A}_2 = -\mathbf{A}_1(\boldsymbol{\omega}\times)\mathbf{A}_2 \tag{6d}$$

It can be easily checked that the attitude kinematics (1) satisfies (5) and therefore, it is group affine. According to Lie group theory [12], the vector error state models corresponding to the group affine model are trajectory-independent. Moreover, if the observations take the following particular forms, the linearized observation model will also be trajectory-independent.

$$\begin{aligned}&\text{Left-Invariant Observation: } \mathbf{y} = \chi\boldsymbol{\mu}\\ &\text{Right-Invariant Observation: } \mathbf{y} = \chi^{-1}\boldsymbol{\mu}\end{aligned} \tag{7}$$

where $\boldsymbol{\mu}$ is a constant vector. It can be checked that the measurement model (3) is left-invariant and the linearized observation model corresponding to the left error definition will be trajectory-independent. However, at certain time instant, both the observation vector $\mathbf{b}_k$ and the reference vector $\mathbf{r}_k$ can be viewed as known values. Therefore, (3) can be rewritten as

$$\mathbf{r}_k = \mathbf{A}(t_k)^{-1}\mathbf{b}_k - \mathbf{A}(t_k)^{-1}\boldsymbol{\upsilon}_k \tag{8}$$

That is to say, the vector observation model can also be viewed as right-invariant. Therefore, the linearized observation model corresponding to the right error definition will also be trajectory-independent.

In next section, we will present the explicit error state-space models for the attitude estimation problem using vector observations and their trajectory-independence will be clearly shown. This is also the starting point of the modification of the traditional MEKF.

## III. Error State-Space Models on $\mathbb{SO}(3)+\mathbb{R}^3$

### A. Error State-Space Model with Body Frame Attitude Error

In the traditional MEKF, the attitude error is defined as

$$\mathbf{A}(\boldsymbol{\delta}\mathbf{q}) = \mathbf{A}(\mathbf{q})\mathbf{A}(\hat{\mathbf{q}})^{-1} \tag{9}$$

Here, we add the quaternion dependence of each attitude matrix because in MEKF framework, the global attitude is propagated in quaternion form. The attitude error in (9) is in the right error form. From the frame transformation perspective, it denotes the attitude error in body frame. This is a little different with those in [11, 16, 20, 21]. The difference is caused by the fact that the global attitude matrix in (1) and (3) denotes attitude transformation from reference frame to body frame while the attitude matrix in [11, 16, 20, 21] denotes attitude transformation from body frame to reference frame.

We term the error in (9) as group error. The corresponding vector form is denoted as $\boldsymbol{\delta\alpha}^b$ . With small attitude error assumption, their relationship is given by

$$\mathbf{A}(\boldsymbol{\delta}\mathbf{q}) = \mathbf{I}_{3\times3} - \left(\boldsymbol{\delta\alpha}^b \times\right) \tag{10}$$

Denote that vector state as $\boldsymbol{\delta}\mathbf{x}_{\text{body}} = \begin{bmatrix} \boldsymbol{\delta\alpha}^{bT} & \boldsymbol{\delta\beta}^T \end{bmatrix}^T$ , where $\boldsymbol{\delta\beta} = \boldsymbol{\beta} - \hat{\boldsymbol{\beta}}$ . The subscript "body" means that the involved attitude error is expressed in body frame. The process model corresponding to (1) and (2) is given by

$$\boldsymbol{\delta}\dot{\mathbf{x}}_{\text{body}} = \mathbf{F}_{\text{body}}\left(\hat{\mathbf{x}}(t), t\right)\boldsymbol{\delta}\mathbf{x}_{\text{body}} + \mathbf{G}_{\text{body}}(t)\mathbf{w}(t) \tag{11a}$$

where

$$\mathbf{F}_{\text{body}}\left(\hat{\mathbf{x}}(t), t\right) = \begin{bmatrix} -(\hat{\boldsymbol{\omega}}\times) & -\mathbf{I}_{3\times3} \\ \mathbf{0}_{3\times3} & \mathbf{0}_{3\times3} \end{bmatrix} \tag{11b}$$

$$\mathbf{G}_{\text{body}}(t) = \begin{bmatrix} -\mathbf{I}_{3\times3} & \mathbf{0}_{3\times3} \\ \mathbf{0}_{3\times3} & \mathbf{I}_{3\times3} \end{bmatrix} \tag{11c}$$

In (11), $\mathbf{w}(t) = \begin{bmatrix} \boldsymbol{\eta}_{\mathbf{v}}^T(t) & \boldsymbol{\eta}_{\mathbf{u}}^T(t) \end{bmatrix}^T$ , $\hat{\boldsymbol{\omega}} = \tilde{\boldsymbol{\omega}} - \hat{\boldsymbol{\beta}}$ . It can be seen from (11) that if the gyroscope bias is not considered, the differential model corresponding to $\boldsymbol{\delta\alpha}^b$ is independent of $\mathbf{A}(\hat{\mathbf{q}})$ . This is consistent with the group affine analysis of (1). Unfortunately, as noted in [12, 18], there is no Lie group that includes the bias terms while also having the dynamics satisfy the group affine property. However, much of the group affine structure of attitude can be preserved even with bias augmentation.

With attitude error definition in (9), the linearized measurement transition matrix corresponding to (3) is given by [19]

$$\mathbf{H}_{\text{body},k}\left(\hat{\mathbf{x}}_k^-\right) = \begin{bmatrix} \left[\mathbf{A}(\hat{\mathbf{q}}_k^-)\mathbf{r}_k \times\right] & \mathbf{0}_{3\times3} \end{bmatrix} \tag{12}$$

It is seemingly that (12) is dependent on the global attitude prediction $\mathbf{A}(\hat{\mathbf{q}}_k^-)$ , which is not consistent with the invariance analysis of the vector observation model. Essentially, this is just the reason why the traditional MEKF performs not so well with large initialization errors in previous works. However, according to (3) and (12), a modified measurement transition matrix can be given by

$$\mathbf{H}_{\text{body},k} = \begin{bmatrix} \left(\tilde{\mathbf{b}}_k \times\right) & \mathbf{0}_{3\times3} \end{bmatrix} \tag{13}$$

where $\tilde{\mathbf{b}}_k$ denotes the measured vector observation with noise. It is now shown that the transition matrix in (13) is trajectory-independent. According to the Lie group theory, it is possible to recover the nonlinear attitude error $\mathbf{A}(\boldsymbol{\delta}\mathbf{q})$ with the linear state-space model (11) and (13). Here we have not claimed that the nonlinear attitude error can be "exactly" recovered from its linear vector error state estimate. This is because that the gyroscope bias augmentation has made the process model not totally trajectory-independent as shown in (11).

*B. Error State-Space Model with Reference Frame Attitude Error*

For MEKF, the attitude error can also be defined as

$$\mathbf{A}\left(\boldsymbol{\delta}\mathbf{q}\right) = \mathbf{A}\left(\hat{\mathbf{q}}\right)^{-1}\mathbf{A}\left(\mathbf{q}\right) \tag{14}$$

The attitude error in (14) is in the left error form. From the frame transformation perspective, it denotes the attitude error in reference (inertial) frame. The MEKF with reference attitude error definition was firstly studied by Gai et al. in [22]. The attitude quaternion in [22] defines the transformation from the body frame to the inertial frame, which is different with that in [4, 5, 9, 19]. With the same attitude definition as in [4, 5, 9, 19], the state-space model for MEKF with reference attitude error definition is re-derived in [17].

Similarly with (10), the attitude error in (14) can also be approximated as

$$\mathbf{A}(\boldsymbol{\delta}\mathbf{q}) = \mathbf{I}_{3\times3} - \left(\boldsymbol{\delta\alpha}^r \times\right) \tag{15}$$

Denote that vector state as $\boldsymbol{\delta}\mathbf{x}_{\text{ref}} = \begin{bmatrix} \boldsymbol{\delta\alpha}^{rT} & \boldsymbol{\delta\beta}^T \end{bmatrix}^T$ , where $\boldsymbol{\delta\beta} = \boldsymbol{\beta} - \hat{\boldsymbol{\beta}}$. The subscript "ref" means that the involved attitude error is expressed in reference frame. The process model of $\boldsymbol{\delta}\mathbf{x}_{\text{ref}}$ is given by

$$\boldsymbol{\delta}\dot{\mathbf{x}}_{\text{ref}} = \mathbf{F}_{\text{ref}}\left(\hat{\mathbf{x}}(t), t\right)\boldsymbol{\delta}\mathbf{x}_{\text{ref}} + \mathbf{G}_{\text{ref}}(t)\mathbf{w}(t) \tag{16a}$$

where

$$\mathbf{F}_{\text{ref}}\left(\hat{\mathbf{x}}(t), t\right) = \begin{bmatrix} \mathbf{0}_{3\times3} & -\mathbf{A}\left(\hat{\mathbf{q}}\right)^T \\ \mathbf{0}_{3\times3} & \mathbf{0}_{3\times3} \end{bmatrix} \tag{16b}$$

$$\mathbf{G}_{\text{ref}}(t) = \begin{bmatrix} -\mathbf{A}\left(\hat{\mathbf{q}}\right)^T & \mathbf{0}_{3\times3} \\ \mathbf{0}_{3\times3} & \mathbf{I}_{3\times3} \end{bmatrix} \tag{16c}$$

It can be seen from (16) that if the gyroscope bias is not considered, the differential model corresponding to $\boldsymbol{\delta\alpha}^r$ is independent of $\mathbf{A}(\hat{\mathbf{q}})$ . This is consistent with the group affine analysis of (1). The augmentation of gyroscope bias introduces model dependence of the global attitude estimate $\mathbf{A}(\hat{\mathbf{q}})$ .

With attitude error definition in (15), the linearized measurement model corresponding to (3) is given by [17]

$$\mathbf{H}_{\text{ref},k}\left(\hat{\mathbf{x}}_k^-\right) = \begin{bmatrix} \left[\left(\mathbf{A}(\hat{\mathbf{q}}_k^-)\mathbf{r}_k\right)\times\right]\mathbf{A}\left(\hat{\mathbf{q}}_k^-\right) & \mathbf{0}_{3\times3} \end{bmatrix} \tag{17}$$

It is shown that the measurement transition matrix in (17) is dependent on the global attitude prediction $\mathbf{A}(\hat{\mathbf{q}}_k^-)$ , which is not consistent with the invariance analysis of the vector observation model. However, (17) can be reorganized as

$$\begin{aligned} \mathbf{H}_{\text{ref},k}(\hat{\mathbf{x}}_k^-) &= [\mathbf{A}(\hat{\mathbf{q}}_k^-)\mathbf{A}(\hat{\mathbf{q}}_k^-)^{-1}[(\mathbf{A}(\hat{\mathbf{q}}_k^-)\mathbf{r}_k)\times]\mathbf{A}(\hat{\mathbf{q}}_k^-) \quad \mathbf{0}_{3\times3}] \\ &= [\mathbf{A}(\hat{\mathbf{q}}_k^-)[(\mathbf{A}(\hat{\mathbf{q}}_k^-)^{-1}\mathbf{A}(\hat{\mathbf{q}}_k^-)\mathbf{r}_k)\times] \quad \mathbf{0}_{3\times3}] \\ &= [\mathbf{A}(\hat{\mathbf{q}}_k^-)(\mathbf{r}_k\times) \quad \mathbf{0}_{3\times3}] \end{aligned} \tag{18}$$

Although (18) is now still dependent on the global attitude, through operation in the following theorem, it is possible to derive a global attitude independent measurement transition matrix.

**Theorem 1** [10]: *Applying a same linear function to the*

*linearized measurement transition matrix and innovation term of an EKF before computing the gains does not change the results of the filter*.

The explicit proof is presented in [10]. Based on **Theorem 1**, the matrix in (18) can be transformed as

$$\mathbf{H}_{\text{ref},k,trans} = \mathbf{A}\left(\hat{\mathbf{q}}_k^-\right)^{-1}\mathbf{H}_k\left(\hat{\mathbf{x}}_k^-\right) = \left[(\mathbf{r}_k\times)\quad \mathbf{0}_{3\times3}\right] \tag{19}$$

Accordingly, the innovation should be transformed as

$$\mathbf{n}_{k,trans} = \mathbf{A}\left(\hat{\mathbf{q}}_k^-\right)^{-1}\left[\mathbf{b}_k - \mathbf{A}\left(\hat{\mathbf{q}}_k^-\right)\mathbf{r}_k\right] = \mathbf{A}\left(\hat{\mathbf{q}}_k^-\right)^{-1}\mathbf{b}_k - \mathbf{r}_k \tag{20}$$

The measurement noise covariance should also be transformed as

$$\mathbf{R}_{k,trans} = \mathbf{A}\left(\hat{\mathbf{q}}_k^-\right)^{-1}\mathbf{R}_k\mathbf{A}\left(\hat{\mathbf{q}}_k^-\right) \tag{21}$$

where $\mathbf{R}_k$ is the covariance corresponding to $\boldsymbol{\upsilon}_k$ in (3). Now, the measurement transition matrix $\mathbf{H}_{\text{ref},k,trans}$ is trajectory independent.

***Remark 1:*** It should be noted that making use of the trajectory-independent matrix (19) and the transformed innovation (20) has the same performance with that making use of (17) or (18) and traditional innovation. The transformation procedures from (18) to (21) are only used to demonstrate that the invariant measurement (3) can result in trajectory-independent measurement transition matrix corresponding to reference attitude error definition.

## IV. Error State-Space Models on $\mathbb{SE}(3)$

In [16], the attitude estimation has also been investigated making use of $\mathbb{SE}(3)$. Through formulating the attitude and gyroscope bias as a compact state on $\mathbb{SE}(3)$, both the GEKF and QRIEKF can be derived. In this section, the explicit error state-space model is directly presented with group state definition on $\mathbb{SE}(3)$.

For spacecraft attitude estimation, the group state on $\mathbb{SE}(3)$ is defined as

$$\chi = \begin{bmatrix} \mathbf{A}(\mathbf{q}) & \boldsymbol{\beta} \\ \mathbf{0}_{1\times3} & 1 \end{bmatrix} \tag{22}$$

It can also be demonstrated that with the group state (22), its dynamic model does not satisfy (5). Therefore, the resulting error state model will be dependent on the global state estimate, which will be shown in following (25) and (28).

### *A. Error State-Space Model with Right Error*

The right group error is corresponding to (9) and it is defined as

$$\delta\chi_{\text{right}} = \chi\hat{\chi}^{-1} = \begin{bmatrix} \mathbf{A}(\mathbf{q})\mathbf{A}(\hat{\mathbf{q}})^{-1} & \boldsymbol{\beta} - \mathbf{A}(\mathbf{q})\mathbf{A}(\hat{\mathbf{q}})^{-1}\hat{\boldsymbol{\beta}} \\ \mathbf{0}_{1\times3} & 1 \end{bmatrix} \tag{23}$$

The vector form corresponding to $\delta\chi_{\text{right}}$ is denoted as $\mathbf{dx}_{\text{right}} = \left[\mathbf{d}\boldsymbol{\alpha}^{bT}\quad \mathbf{d}\boldsymbol{\beta}^{bT}\right]^T$.The subscript "right" means that the error state is corresponding to the right group error. $\mathbf{d}\boldsymbol{\alpha}^b$ is corresponding to the attitude error $\mathbf{A}(\mathbf{q})\mathbf{A}(\hat{\mathbf{q}})^{-1}$. It is clearly shown that $\mathbf{d}\boldsymbol{\alpha}^b$ is just the attitude error in (9). $\mathbf{d}\boldsymbol{\beta}^b$ is corresponding to the gyroscope drift bias error and is given by

$$\begin{aligned}\mathbf{d}\boldsymbol{\beta}^b &= \boldsymbol{\beta} - \mathbf{A}(\mathbf{q})\mathbf{A}(\hat{\mathbf{q}})^{-1}\hat{\boldsymbol{\beta}} \\ &= \boldsymbol{\beta} - \left[\mathbf{I}_{3\times3} - (\mathbf{d}\boldsymbol{\alpha}\times)\right]\hat{\boldsymbol{\beta}} = \boldsymbol{\delta\beta} - \left(\hat{\boldsymbol{\beta}}\times\right)\mathbf{d}\boldsymbol{\alpha}\end{aligned} \tag{24}$$

where we have made the same approximation for $\mathbf{d}\boldsymbol{\alpha}^b$ as (10). The process model of $\mathbf{dx}_{\text{right}}$ is given by

$$\mathbf{d}\dot{\mathbf{x}}_{\text{right}} = \mathbf{F}_{\text{right}}\left(\hat{\chi}(t),t\right)\mathbf{dx}_{\text{right}} + \mathbf{G}_{\text{right}}\left(\hat{\chi}(t),t\right)\mathbf{w}(t) \tag{25a}$$

where

$$\mathbf{F}_{\text{right}}\left(\hat{\chi}(t),t\right) = \begin{bmatrix} -(\tilde{\boldsymbol{\omega}}\times) & -\mathbf{I}_{3\times3} \\ \left(\hat{\boldsymbol{\beta}}\times\right)(\tilde{\boldsymbol{\omega}}\times) & \left(\hat{\boldsymbol{\beta}}\times\right) \end{bmatrix} \tag{25b}$$

$$\mathbf{G}_{\text{right}}\left(\hat{\chi}(t),t\right) = \begin{bmatrix} -\mathbf{I}_{3\times3} & \mathbf{0}_{3\times3} \\ \left(\hat{\boldsymbol{\beta}}\times\right) & \mathbf{I}_{3\times3} \end{bmatrix} \tag{25c}$$

It is clearly shown that (25b) is just the process transition matrix used in GEKF. The defined error state $\mathbf{dx}_{\text{right}}$ is a little different with that in [17]. This is because that we intend to derive the consistent result with GEKF. It is shown in (25) that with the right error definition, the process transition matrix is now dependent on the global state estimate $\hat{\boldsymbol{\beta}}$. This is also the result of the fact that the dynamic model of the group state (22) is not group affine.

Since the attitude error element of $\mathbf{dx}_{\text{right}}$ is with the same definition with that of $\boldsymbol{\delta}\mathbf{x}_{\text{body}}$, the measurement transition matrix corresponding to $\mathbf{dx}_{\text{right}}$ is the same with (12) or (13). Based on the invariance theory, (13) will be more preferred than (12).

### *B. Error State-Space Model with Left Error*

The left group error is corresponding to (14) and it is defined as

$$\delta\chi_{\text{left}} = \hat{\chi}^{-1}\chi = \begin{bmatrix} \mathbf{A}(\hat{\mathbf{q}})^{-1}\mathbf{A}(\mathbf{q}) & \mathbf{A}(\hat{\mathbf{q}})^{-1}\left(\boldsymbol{\beta} - \hat{\boldsymbol{\beta}}\right) \\ \mathbf{0}_{1,3} & 1 \end{bmatrix} \tag{26}$$

The vector form corresponding to $\delta\chi_{\text{left}}$ is denoted as $\mathbf{dx}_{\text{left}} = \left[\mathbf{d}\boldsymbol{\alpha}^{rT}\quad \mathbf{d}\boldsymbol{\beta}^{rT}\right]^T$.The subscript "left" means that the error state is corresponding to the left group error. $\mathbf{d}\boldsymbol{\alpha}^r$ is corresponding to the attitude error $\mathbf{A}(\hat{\mathbf{q}})^{-1}\mathbf{A}(\mathbf{q})$. It is clearly shown that $\mathbf{d}\boldsymbol{\alpha}^r$ is just the attitude error in (14). $\mathbf{d}\boldsymbol{\beta}^r$ is corresponding to the gyroscope bias error and is given by

$$\mathbf{d}\boldsymbol{\beta}^r = \mathbf{A}(\hat{\mathbf{q}})^{-1}\left(\boldsymbol{\beta} - \hat{\boldsymbol{\beta}}\right) = \mathbf{A}(\hat{\mathbf{q}})^{-1}\boldsymbol{\delta\beta} \tag{27}$$

From the attitude transformation perspective, (27) can be viewed as transforming the gyroscope bias error from body frame to reference frame. This is also the insight in [16] that "*the group actions can be intuitively viewed as some coordinate transformations*".

The process model of $\mathbf{dx}_{\text{left}}$ is given by

$$\mathbf{d}\dot{\mathbf{x}}_{\text{left}} = \mathbf{F}_{\text{left}}\left(\hat{\chi}(t),t\right)\mathbf{dx}_{\text{left}} + \mathbf{G}_{\text{left}}\left(\hat{\chi}(t),t\right)\mathbf{w}(t) \tag{28a}$$

where

$$\mathbf{F}_{\text{left}}\left(\hat{\chi}(t),t\right)=\begin{bmatrix}\mathbf{0}_{3\times3} & -\mathbf{I}_{3\times3}\\ \mathbf{0}_{3\times3} & \left[\left(\mathbf{A}\left(\hat{\mathbf{q}}\right)^T\hat{\boldsymbol{\omega}}\right)\times\right]\end{bmatrix} \quad (28b)$$

$$\mathbf{G}_{\text{left}}\left(\hat{\chi}(t),t\right)=\begin{bmatrix}-\mathbf{A}\left(\hat{\mathbf{q}}\right)^T & \mathbf{0}_{3\times3}\\ \mathbf{0}_{3\times3} & \mathbf{A}\left(\hat{\mathbf{q}}\right)^T\end{bmatrix} \quad (28c)$$

It is clearly shown that (28b) is just the process transition matrix used in QRIEKF in [16]. The defined error state $\mathbf{dx}_{\text{left}}$ is a little different with that in [17]. This is because that we intend to derive the consistent result with QRIEKF. In this paper, the process model with form (28) is corresponding to left group state error while such form is corresponding to right group state error in [16]. This is also caused by the difference between the meanings of the attitude matrix. Specially, the attitude matrix in this paper denotes attitude transformation from reference frame to body frame while the attitude matrix in [16] denotes attitude transformation from body frame to reference frame. It is shown that (28b) is also dependent on the global state estimate. Compared with (16), the dependence has been moved from attitude error model to gyroscope bias error model.

Since the attitude error element of $\mathbf{dx}_{\text{left}}$ is with the same definition with that of $\boldsymbol{\delta}\mathbf{x}_{\text{ref}}$, the measurement transition matrix corresponding to $\mathbf{dx}_{\text{left}}$ is the same with (17) or (19).

***Remark 2:*** In QRIEKF, the measurement transition matrix (19) is used. In [16], such trajectory-independent measurement transition matrix is derived from the invariance theory. In this paper, we presented another perspective on the derivation of the trajectory-independent measurement transition matrix as shown in (17)-(19).

## V. Application Aspects

The explicit MEKF procedures are summarized in Table 1. The transition matrices $\mathbf{F}(t)$, $\mathbf{G}(t)$ and $\mathbf{H}_k$ in Table I. take different forms as in (11)-(13), (16)-(19), (25) and (28), according to different error state definitions. The GEKF and QRIEKF have also been viewed as type of MEKF. The essential difference between different algorithms lies in the error state definition. Different error state definitions result in different state-space models and different attitude update procedures.

Table 1 MEKF for Attitude Estimation Using Vector Observations

| | |
|---|---|
| **Initialize** | $\mathbf{q}(t_0)=\hat{\mathbf{q}}_0$, $\boldsymbol{\beta}(t_0)=\hat{\boldsymbol{\beta}}_0$, $\mathbf{P}(t_0)=\mathbf{P}_0$ |
| **Gain** | $\mathbf{K}_k=\mathbf{P}_k^-\mathbf{H}_k^T\left[\mathbf{H}_k\mathbf{P}_k^-\mathbf{H}_k^T+\mathbf{R}_k\right]^{-1}$ |
| | $\mathbf{P}_k^+=\left[\mathbf{I}_{6\times6}-\mathbf{K}_k\mathbf{H}_k\right]\mathbf{P}_k^-$ |
| **Update** | $\Delta\hat{\mathbf{x}}_k=\mathbf{K}_k\left[\mathbf{b}_k-\mathbf{A}\left(\hat{\mathbf{q}}_k^-\right)\mathbf{r}_k\right]$ |
| | $\hat{\mathbf{x}}_k^+=\hat{\mathbf{x}}_k^-\boxplus\Delta\hat{\mathbf{x}}_k$ |
| | $\hat{\boldsymbol{\omega}}(t)=\tilde{\boldsymbol{\omega}}(t)-\hat{\boldsymbol{\beta}}(t)$ |
| **Propagation** | $\dot{\hat{\mathbf{q}}}(t)=\frac{1}{2}\boldsymbol{\Xi}\left(\hat{\mathbf{q}}(t)\right)\hat{\boldsymbol{\omega}}(t)$ |
| | $\dot{\mathbf{P}}(t)=\mathbf{F}(t)\mathbf{P}(t)+\mathbf{P}(t)\mathbf{F}^T(t)+\mathbf{G}(t)\mathbf{Q}(t)\mathbf{G}^T(t)$ |

The retraction operation $\boxplus$ denotes how to update the global state using the local error state estimate. Specially, for error state definition $\boldsymbol{\delta}\mathbf{x}_{\text{body}}$, $\Delta\hat{\mathbf{x}}_k=\begin{bmatrix}\boldsymbol{\delta}\hat{\boldsymbol{\alpha}}_k^{b+T} & \boldsymbol{\delta}\hat{\boldsymbol{\beta}}_k^{+T}\end{bmatrix}^T$, the retraction operation is given by

$$\hat{\mathbf{q}}_k^+=\exp_{\mathbf{q}}\left(\boldsymbol{\delta}\hat{\boldsymbol{\alpha}}_k^{b+}\right)\otimes\hat{\mathbf{q}}_k^- \quad (29a)$$

$$\hat{\boldsymbol{\beta}}_k^+=\hat{\boldsymbol{\beta}}_k^-+\boldsymbol{\delta}\hat{\boldsymbol{\beta}}_k^+ \quad (29b)$$

where

$$\exp_{\mathbf{q}}\left(\boldsymbol{\delta\alpha}\right)=\begin{bmatrix}\dfrac{\boldsymbol{\delta\alpha}}{\left\|\boldsymbol{\delta\alpha}\right\|}\sin\left(\left\|\boldsymbol{\delta\alpha}\right\|/2\right)\\ \cos\left(\left\|\boldsymbol{\delta\alpha}\right\|/2\right)\end{bmatrix} \quad (30)$$

For error state definition $\boldsymbol{\delta}\mathbf{x}_{\text{ref}}$, $\Delta\hat{\mathbf{x}}_k=\begin{bmatrix}\boldsymbol{\delta}\hat{\boldsymbol{\alpha}}_k^{r+T} & \boldsymbol{\delta}\hat{\boldsymbol{\beta}}_k^{+T}\end{bmatrix}^T$, the retraction operation is given by

$$\hat{\mathbf{q}}_k^+=\hat{\mathbf{q}}_k^-\otimes\exp_{\mathbf{q}}\left(\boldsymbol{\delta}\hat{\boldsymbol{\alpha}}_k^{r+}\right) \quad (31a)$$

$$\hat{\boldsymbol{\beta}}_k^+=\hat{\boldsymbol{\beta}}_k^-+\boldsymbol{\delta}\hat{\boldsymbol{\beta}}_k^+ \quad (31b)$$

For error state definition $\mathbf{dx}_{\text{right}}$, $\Delta\hat{\mathbf{x}}_k=\begin{bmatrix}\mathbf{d}\hat{\boldsymbol{\alpha}}_k^{b+T} & \mathbf{d}\hat{\boldsymbol{\beta}}_k^{b+T}\end{bmatrix}^T$, the retraction operation is given by

$$\hat{\mathbf{q}}_k^+=\exp_{\mathbf{q}}\left(\mathbf{d}\hat{\boldsymbol{\alpha}}_k^{b+}\right)\otimes\hat{\mathbf{q}}_k^- \quad (32a)$$

$$\hat{\boldsymbol{\beta}}_k^+=\hat{\boldsymbol{\beta}}_k^-+\mathbf{d}\hat{\boldsymbol{\beta}}_k^{b+}+\left(\hat{\boldsymbol{\beta}}_k^-\times\right)\mathbf{d}\hat{\boldsymbol{\alpha}}_k^{b+} \quad (32b)$$

For error state definition $\mathbf{dx}_{\text{left}}$, $\Delta\hat{\mathbf{x}}_k=\begin{bmatrix}\mathbf{d}\hat{\boldsymbol{\alpha}}_k^{r+T} & \mathbf{d}\hat{\boldsymbol{\beta}}_k^{r+T}\end{bmatrix}^T$, the retraction operation is given by

$$\hat{\mathbf{q}}_k^+=\hat{\mathbf{q}}_k^-\otimes\exp_{\mathbf{q}}\left(\mathbf{d}\hat{\boldsymbol{\alpha}}_k^{r+}\right) \quad (33a)$$

$$\hat{\boldsymbol{\beta}}_k^+=\hat{\boldsymbol{\beta}}_k^-+\mathbf{A}(\hat{\mathbf{q}}_k^-)\mathbf{d}\hat{\boldsymbol{\beta}}_k^{r+} \quad (33b)$$

It should be noted that for error state $\boldsymbol{\delta}\mathbf{x}_{\text{ref}}$ and $\mathbf{dx}_{\text{left}}$, if making use of the global state-independent transition matrix in (19), the transformed innovation in (20) and measurement noise covariance in (21) should be used accordingly.

It should be noted that in the aforementioned methods, the covariance update has not been incorporated for the attitude update. One can refer to [23] for the corresponding covariance update methods.

## VI. Numerical Example

In this section, MEKF with different state error definitions and different state-space models are evaluated via simulated spacecraft attitude estimation using vector observations. Specially, the following five algorithms are evaluated and compared.

The traditional MEKF, that is with state-space model (11) and (12). In this algorithm, the rigorous error quaternion construction (30) has been used. In this respect, here the MEKF is just the QLIEKF in [16].

MEKF with state-space model (11) and (13), denoted as invariant MEKF (IMEKF). Here we make use of the terminology “invariant” to express that the measurement transition matrix is independent of the global state estimate.

MEKF with state-space model (25) and (13), denoted as invariant GEKF (IGEKF). This state-space model with (25) and (12) is just the same with that in GEKF. So, we have made use of the same terminology "GEKF" in [9].

MEKF with state-space model (16) and (19), denoted as MEKF-ref. In this algorithm, the attitude error is expressed in reference frame.

MEKF with state-space model (28) and (19), denoted as QRIEKF. In this algorithm, the attitude error is expressed in reference frame. This is just the filter developed in [16], so we make use of the same terminology.

The used numerical example is the same with that in [16]. In this example, a freely tumbling rigid spacecraft in a circular orbit is considered with altitude of 500 km, inclination angle of $60^\circ$, right ascension of the ascending node of $120^\circ$, argument perigee of $0^\circ$, and initial true anomaly of $0^\circ$ at the epoch of 12:00 Coordinated Universal Time, 1 June 2015. The inertia of the spacecraft is set as $\mathbf{J} = \mathrm{diag}\left[60, 53, 70\right] kg/m^2$. The spacecraft angular velocity is governed by the Euler dynamics

$$\mathbf{J}\dot{\boldsymbol{\omega}} + \boldsymbol{\omega} \times \mathbf{J}\boldsymbol{\omega} = 3\mu \mathbf{r}_{sc} \times \mathbf{J}\mathbf{r}_{sc} / \left\|\mathbf{r}_{sc}\right\|^5 \quad (34)$$

where $\mu$=398600.4418 $kg^3/s^2$ is the Earth's gravitational parameter and $\mathbf{r}_{sc}$ is the spacecraft position vector relative to the Earth center expressed in body frame.

The spacecraft angular velocity is measured by the gyroscopes at a frequency of 0.1 Hz. The vector observations are provided by the sun sensor and magnetometer. The sampling frequency of the two sensors is 1 Hz. The 12th Generation (IGRF-12) model is used to provide the reference magnetic field vector.

The performance of the evaluated filters is quantized and compared using the root-mean-square errors (RMSEs) as

$$\left\|\Delta\boldsymbol{\alpha}\right\|_{\mathrm{RMSE}} = \sqrt{\frac{1}{N_{MC}} \sum_{j=1}^{N_{MC}} \left\|\Delta\boldsymbol{\alpha}^j\right\|} \quad (35)$$

$$\left\|\Delta\boldsymbol{\beta}\right\|_{\mathrm{RMSE}} = \sqrt{\frac{1}{N_{MC}} \sum_{j=1}^{N_{MC}} \left\|\Delta\boldsymbol{\beta}^j\right\|} \quad (36)$$

where $N_{MC}$ is the total Monte Carlo runs. $\Delta\boldsymbol{\alpha}^j$ and $\Delta\boldsymbol{\beta}^j$ are the attitude estimation error and gyroscope bias estimation error respectively for the $j$th Monte Carlo run. It should be noted that although the gyroscope bias error is with the different forms for $\mathbb{SO}(3)+\mathbb{R}^3$ and $\mathbb{SE}(3)$ formulations, the gyroscope bias estimation error $\Delta\boldsymbol{\beta}^j$ is defined with the same form for the evaluated filters as $\Delta\boldsymbol{\beta}^j = \hat{\boldsymbol{\beta}}^j - \boldsymbol{\beta}^j$. This is also the case for the attitude error definition as $\Delta\boldsymbol{\alpha}^j = \hat{\boldsymbol{\alpha}}^j - \boldsymbol{\alpha}^j$, where $\boldsymbol{\alpha}^j$ is the true attitude in Euler angles form and $\hat{\boldsymbol{\alpha}}^j$ is its corresponding estimate by different filters.

*A. Simulations with Large Initial Estimation Errors*

Firstly, scenario with large initial estimation errors is considered. The noise parameters for the gyroscope measurements are given by $\boldsymbol{\eta}_{\mathbf{u}} = \sqrt{10} \times 10^{-10}$ and $\boldsymbol{\eta}_{\mathbf{v}} = \sqrt{10} \times 10^{-7}$. The covariance of sun sensor noise is set to $0.0175^2\mathbf{I}_{3\times3}$ and the covariance of magnetometer noise is set to $0.0873^2\mathbf{I}_{3\times3}$. The true initial attitude and gyroscope bias are generated randomly according to $\boldsymbol{\alpha}(0) \sim N\left(\mathbf{0}_{3\times1}, (150^\circ)^2\mathbf{I}_{3\times3}\right)$ and $\boldsymbol{\beta}(0) \sim N\left(\mathbf{0}_{3\times1}, (20^\circ/h)^2\mathbf{I}_{3\times3}\right)$ .For each filter, the initial attitude error covariance is set as $(150^\circ)^2$ for each axis and the initial bias error covariance is set as $(20^\circ/h)^2$.The initial bias estimate is set to zeros. The initial attitude quaternion estimate is set to unit quaternion. A total of 100 Monte Carlo runs are conducted, each of which spans 60 minutes. The root mean squared errors (RMSEs) of the attitude and gyroscope bias estimates by different filters are shown in Fig. 1 and 2, respectively. It is shown that all the filters with trajectory-independent measurement models have a good performance in terms of both convergence speed and steady-state estimation accuracy. All of them demonstrate significantly better performance than the traditional MEKF and GEKF. In this simulation scenario, the superiority of GEKF over MEKF has not been observed. It can be say that the positive effect provided by the common frame treatment of the gyroscope bias has been counteracted by the negative effect by the trajectory-dependent measurement model. According to the difference between MEKF and IMEKF, it can be concluded that the state prediction-dependent measurement model in MEKF has a much negative effect on the filtering performance. For the four filters with trajectory-independent measurement models, their process models are all trajectory-dependent due to the incorporation of the gyroscope bias as state element. However, regarding the attitude part, the corresponding model is trajectory-independent. This is just the primary reason why these filters can still work well for large initialization errors with the linear Kalman filtering framework. There are slight differences in the steady-state estimates between the four filters with trajectory-independent measurement models. MEKF-ref and QRIEKF perform a little better than IMEKF and IGEKF. Such difference is caused by the incorporation form of the gyroscope drift bias estimate in the process models. For MEKF-ref and QRIEKF, the relation between the attitude error and gyroscope bias error is linear as shown in (16) and (28) while this is nonlinear for IMEKF and IGEKF as shown in (11) and (25). However, the performance of MEKF-ref and QRIEKF is of the same grade. So it cannot be concluded which one of $\mathbb{SE}(3)$ and $\mathbb{SO}(3)$ is better than the other. In previous works, the $\mathbb{SE}(3)$ state formulation is always argued to be better than the $\mathbb{SO}(3)$ formulation. This is because that with $\mathbb{SE}(3)$ formulation, the resultant state-space model is trajectory-independent, while this is not the case for $\mathbb{SO}(3)$ formulation. However, for the spacecraft attitude estimate problem, no matter the $\mathbb{SE}(3)$ or $\mathbb{SO}(3)$ formulation, the attitude error models are both trajectory-independent if the gyroscope bias is not considered. Although IGEKF performs better than IMEKF for attitude estimation in this case, IMEKF performs better than IGEKF in next case. Therefore, we can also not judge which one is superior or inferior.

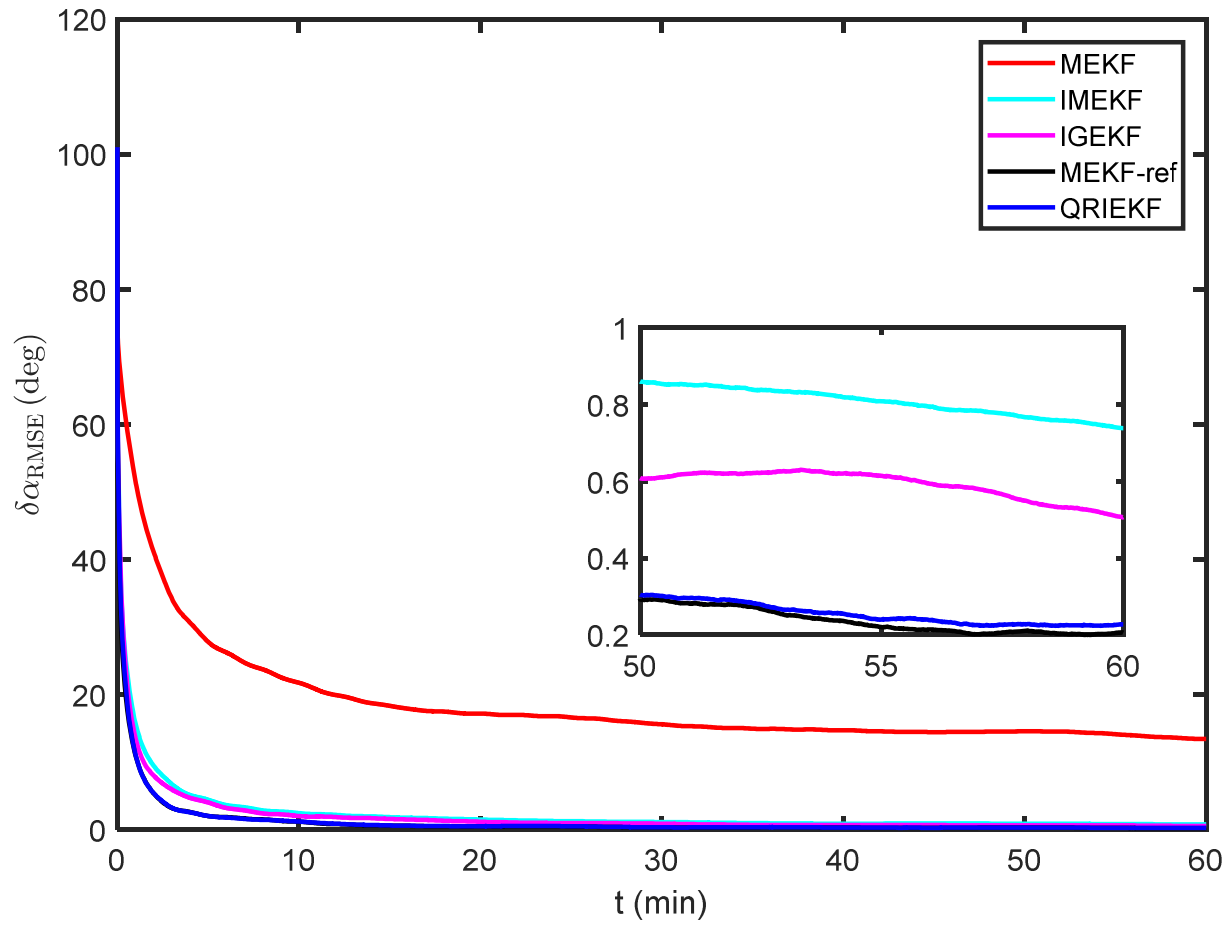

Fig. 1 RMSEs of attitude estimate errors for large initial estimation errors.

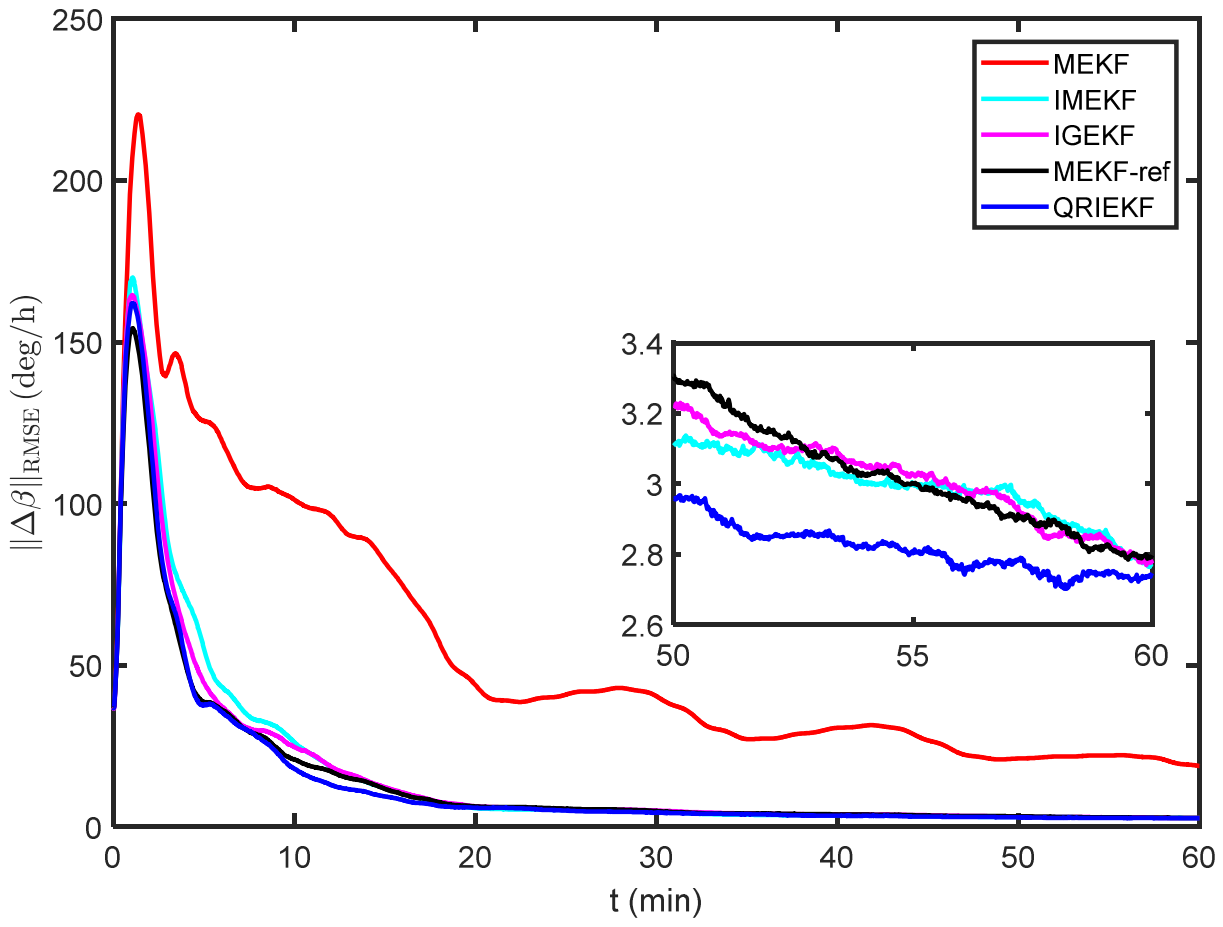

Fig. 2 RMSEs of gyroscope drift bias estimate errors for large initial estimation errors.

## *B. Simulations with a Severe Initial Condition*

Secondly, the severe case with a small initial error covariance but extremely large initial state error is considered. The noise parameters for the gyroscope measurements are given by $\boldsymbol{\eta}_{\mathbf{u}} = \sqrt{10} \times 10^{-8}$ and $\boldsymbol{\eta}_{\mathbf{v}} = \sqrt{10} \times 10^{-5}$. The covariance of sun sensor noise and magnetometer noise are the same with the first case. The initial attitude quaternion estimate is still set to unit quaternion and the initial bias estimate is still set to zeros. But the initial true gyroscope bias is set as $\boldsymbol{\beta}(0) = [100,10,10]^{\circ}/h$ and the initial true attitude quaternion is set as $\mathbf{q}(0) = [1,0,0,0]^T$, which means that the initial attitude error is $180^{\circ}$. For each filter, the initial attitude error covariance is set as $(10^{\circ})^2$ for each axis and the initial bias error covariance is set as $(5^{\circ}/h)^2$. A total of 100 Monte Carlo runs are conducted, each of which spans 80 minutes. The RMSEs of the attitude and gyroscope bias estimates by these filters are shown in Fig. 3 and 4, respectively. The results are consistent with those in last case. All the filters with trajectory-independent measurement models perform much better than the traditional MEKF. This further demonstrates the importance of the trajectory-independent measurement models. In this case, MEKF-ref and QRIEKF still perform a little better than IMEKF and IGEKF for attitude estimation. However, for gyroscope bias estimation, IMEKF performs the best. For attitude estimation, IMEKF performs a litter better than IGEKF, which is opposite with last case. This further indicates that $\mathbb{SE}(3)$ or $\mathbb{SO}(3)$ are only two different state formulations and there is no definitive conclusion that which one is better. Their performance in filtering is problem-dependent.

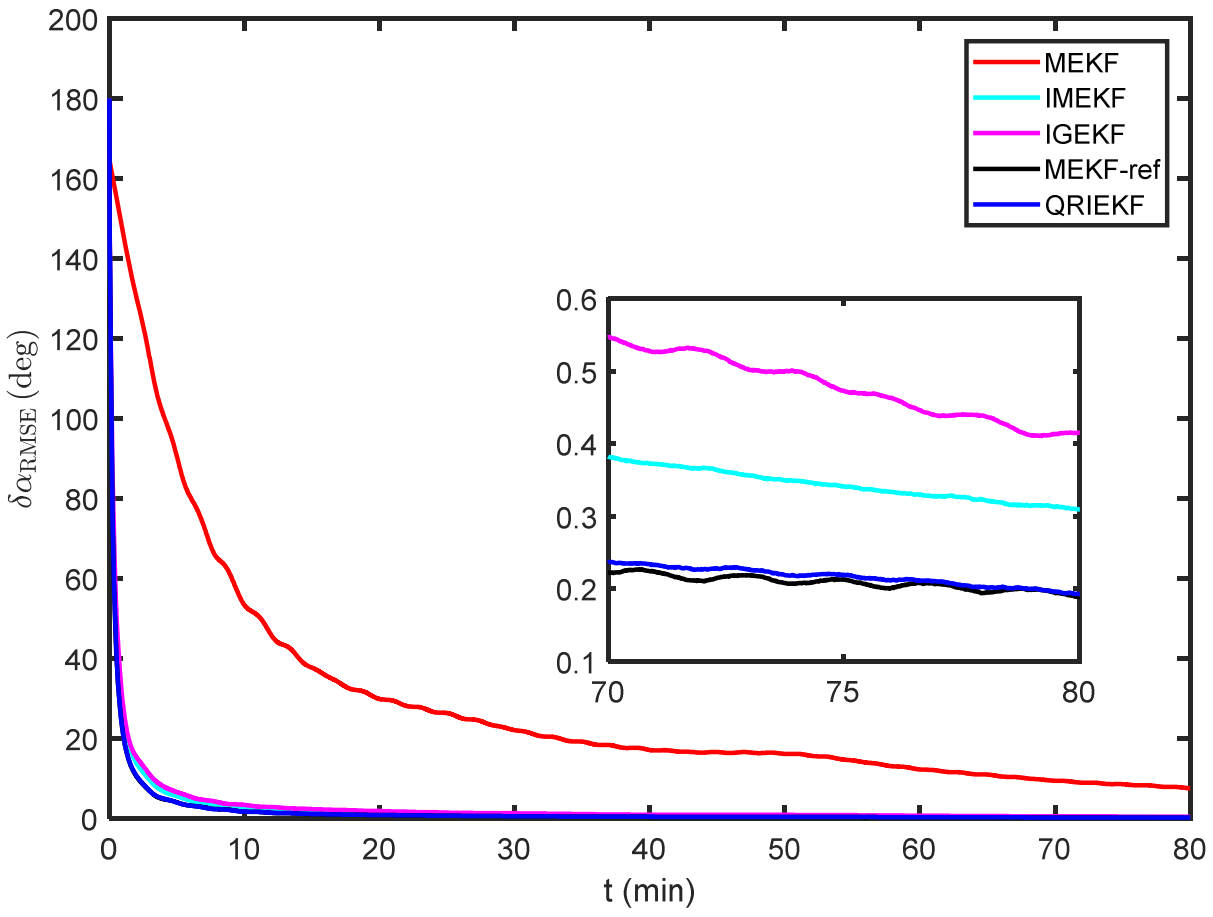

Fig. 3 RMSEs of attitude estimate errors for a severe initial condition.

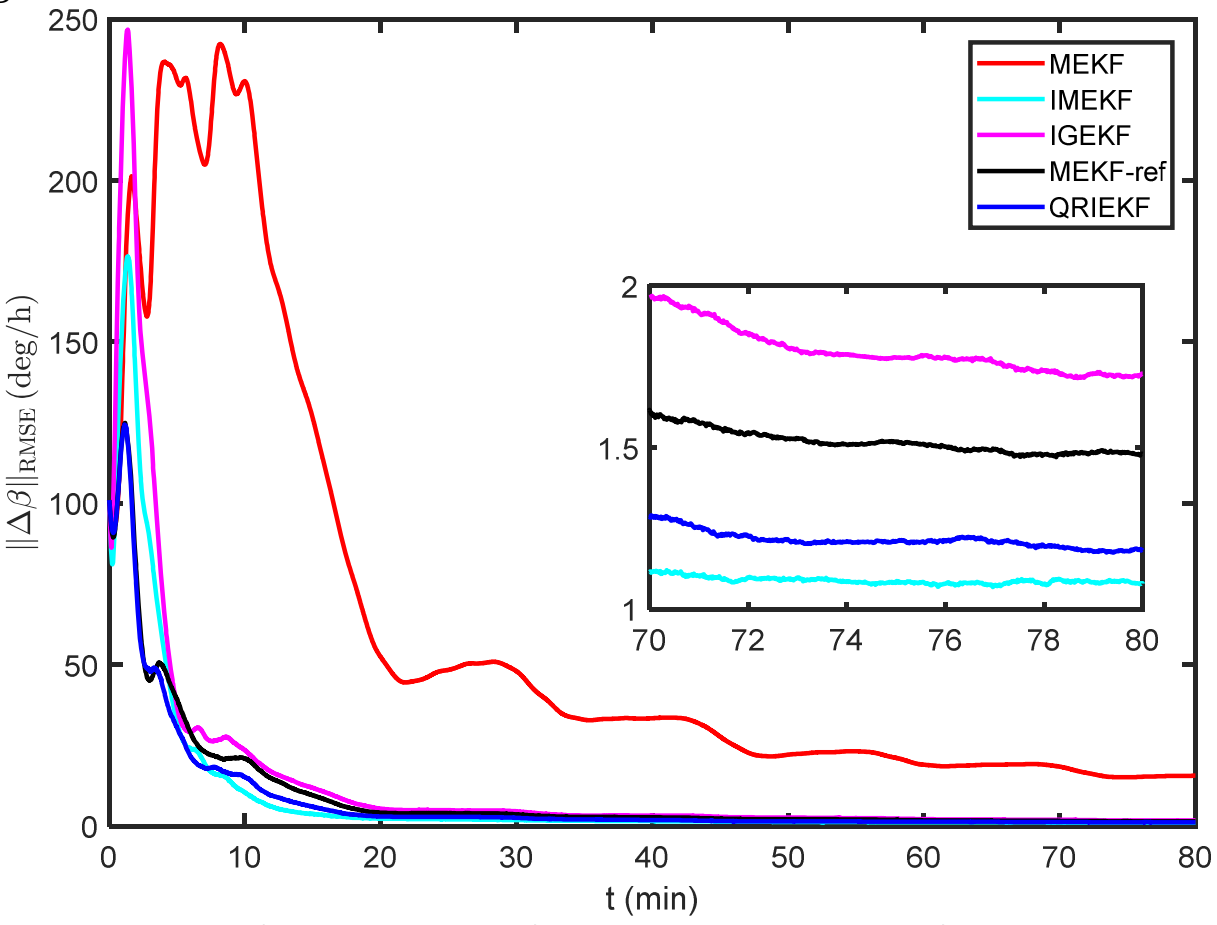

Fig. 4 RMSEs of gyroscope drift bias estimate errors for a severe initial condition.

# VII. Field Test

In order to further demonstrate the superiority of the IMEKF over traditional MEKF, a car-mounted experiment was carried out to collect measurements of ring laser inertial navigation system (INS). The experiment sequence was carried out as follows: after the INS is powered-on, make the car to be static for about 20 minutes. During this stage, the traditional two-stage initial alignment method is carried out to derive the accurate initial attitude for the following swaying stage. Then some peoples get on and off the car frequently to enhance the disturbance artificially and a test segment of about 1000-seconds test data is collected to test the attitude estimation based initial alignment methods.

These measurements provided by the inertial measurement unit are used to construct the vector observations for the attitude estimation based initial alignment. The vector observations for the attitude estimation based initial alignment

are given by [24]

$$\boldsymbol{\alpha}(t) = \mathbf{A}(t)\boldsymbol{\beta}(t) \tag{37}$$

where

$$\boldsymbol{\alpha}(t) = \int_{t_m}^{t} \mathbf{C}_{b(\tau)}^{b(t)} \mathbf{f}^b(\tau)\, d\tau \tag{38}$$

$$\boldsymbol{\beta}(t) = -\int_{t_m}^{t} \mathbf{C}_{n(\tau)}^{i} \mathbf{g}^n\, d\tau \tag{39}$$

where $\mathbf{f}^b$ is the measurement of accelerometer and $\mathbf{g}^n$ is the gravity vector. $\mathbf{C}_{b(\tau)}^{b(t)}$ denotes the attitude transformation from frame $b(\tau)$ to $b(t)$. Similarly explanation can be obtained for $\mathbf{C}_{n(\tau)}^{i}$ . $\Delta t = t - t_m$ denotes the integration interval. In this experiment, it is set as $\Delta t = 10s$ . The detailed calculation procedures for these vector observations can be referred in [24].

For INS initial alignment under swaying conditions, the gyroscope bias cannot be estimated accurately due to its weak observability [25]. With this consideration, the gyroscope bias has not been incorporated into the state. Therefore, there is no need to evaluate IGEKF and QRIEKF. Meanwhile, the novel algorithm proposed in this paper is the IMEKF. So we only evaluate IMEKF and MEKF in this experiment for clarity. The initial attitude provided by the two-stage initial alignment using the static data is added different misalignments to evaluate the ability of handling the large misalignments by the two algorithms. Specially, $10^\circ$ misalignment is added in the initial pitch and roll angles. Different misalignments ranging from $30^\circ$ to $170^\circ$ are added in the initial yaw angle. The alignment results by the two algorithms with different yaw misalignments are shown in Fig. 5 and 6. It is clearly shown that the IMEKF can converge very fast even with very large misalignments. Actually, for this experiment, the convergent speed does even not affected by the different misalignments. In other word, it can be applied ignoring the initial conditions. In contrast, the MEKF can no longer converge when the misalignment becomes large. Moreover, even if it can converge, the steady-state error is also very large.

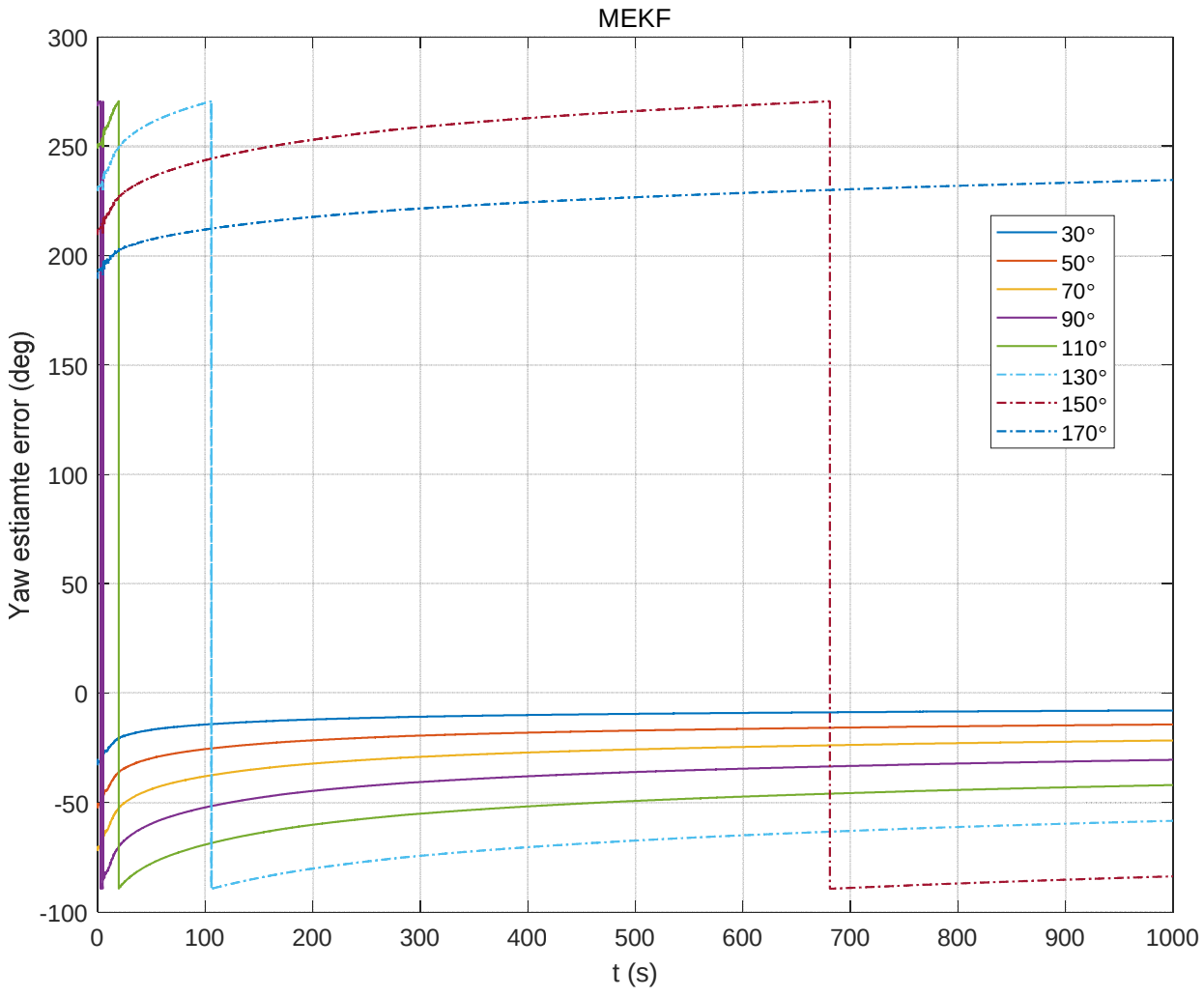

Fig. 5 Yaw angle estimate errors by MEKF with different yaw misalignments

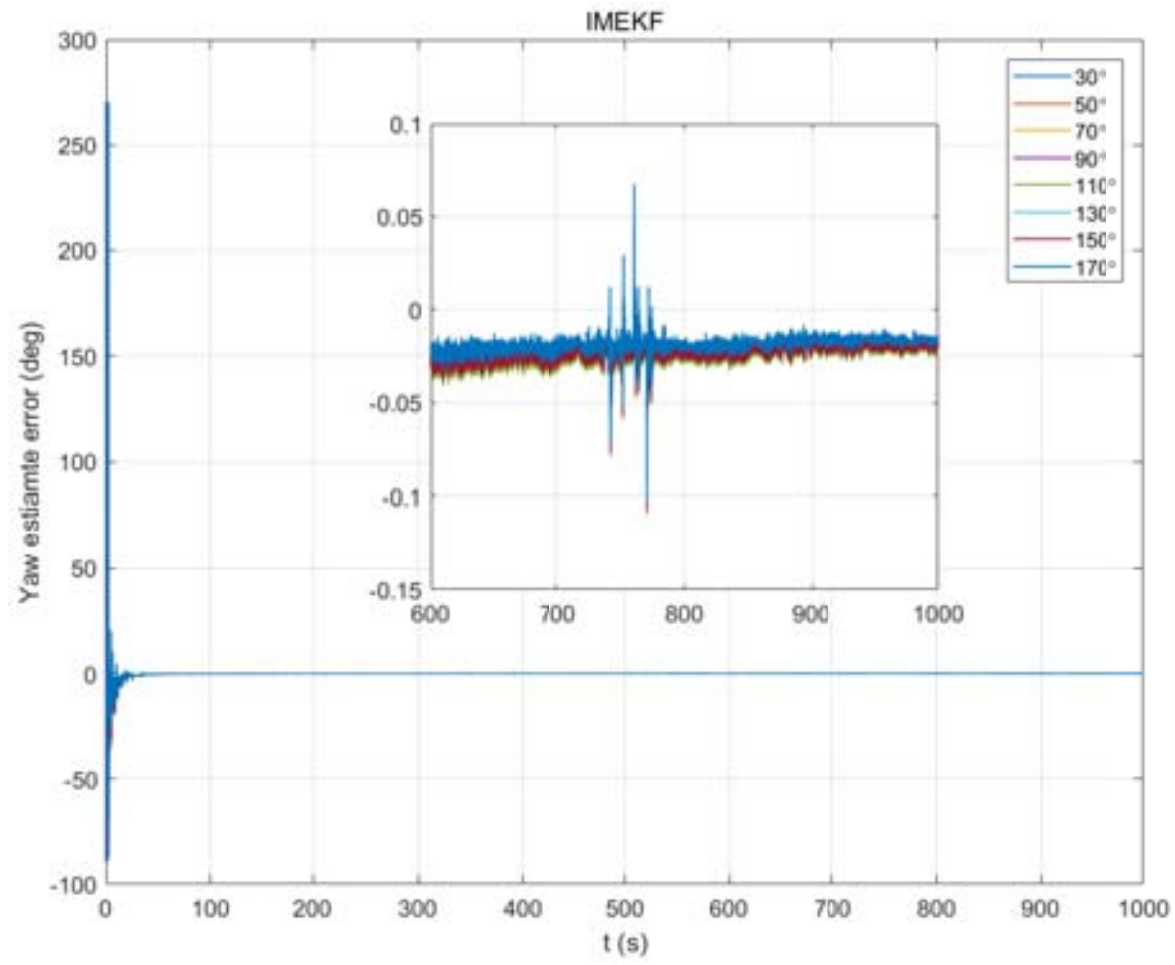

Fig. 6 Yaw angle estimate errors by IMEKF with different yaw misalignments

Specially, with initial misalignment $\left[10^\circ\ 10^\circ\ 170^\circ\right]$, the pith and roll angles estimate results by the two algorithms are shown in Fig. 7 and 8. From Fig. 7 and 8, the swaying condition can be clearly observed. It is shown that, both the two algorithms can converge. However, the steady-state estimate by IMEKF is more accurate than that by MEKF.

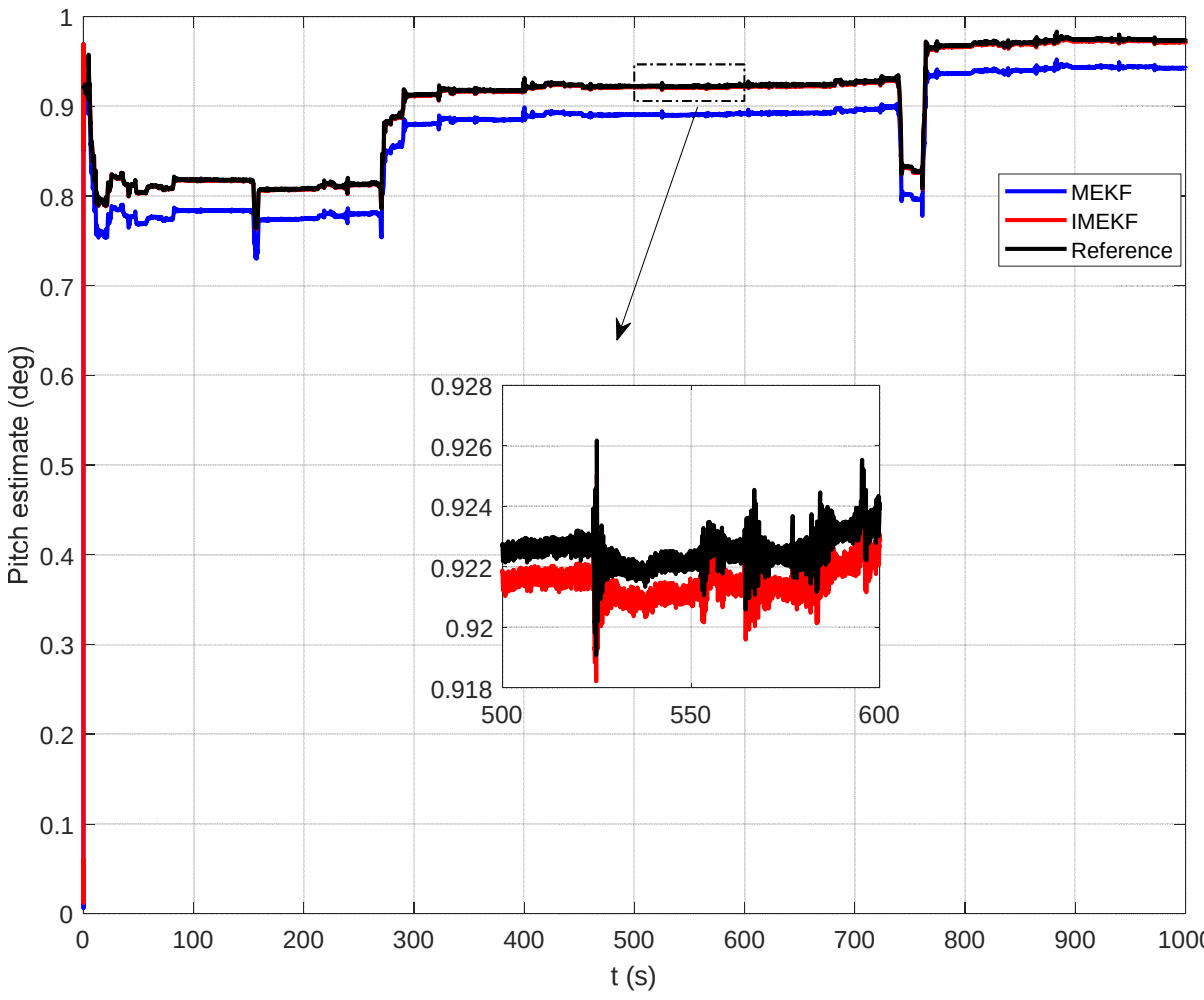

Fig. 7 Pith angle estimate by MEKF and IMEKF

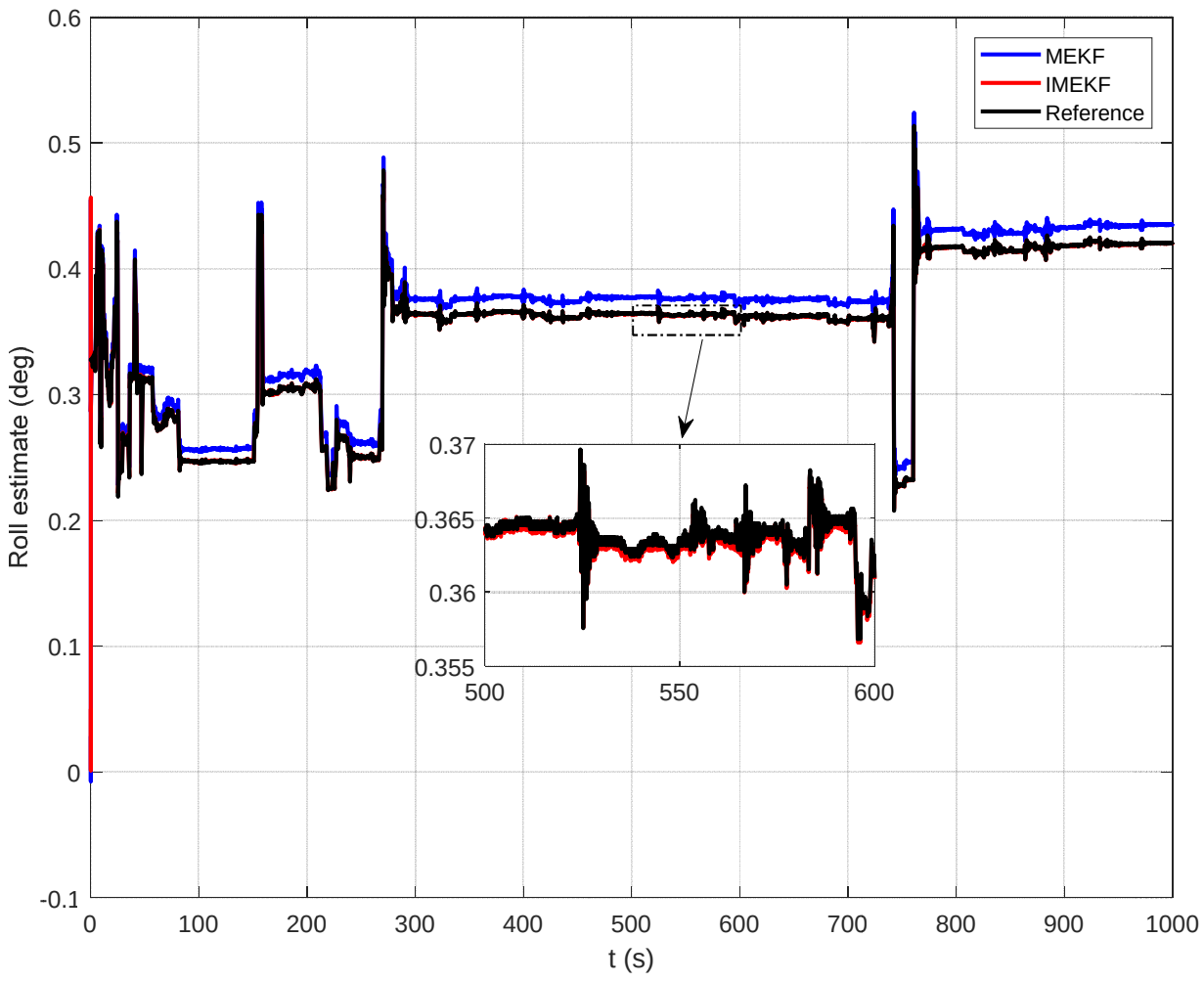

Fig. 8 Roll angle estimate by MEKF and IMEKF

## VIII. Conclusions

In this paper, the MEKF is revisited for spacecraft attitude estimation making use of group Lie theory. It is pointed out that the dynamic model of the spacecraft attitude estimation is group affine and therefore, the corresponding error state model will be trajectory-independent. With the trajectory-independent linear state-space models, it is possible to cope with the large initialization errors even within the linear Kalman filtering framework. However, the used measurement model in existing MEKF is dependent on the attitude prediction, which is inconsistent with invariant theory. Substituting the attitude prediction related vectors with the vector observations in body frame, the resultant measurement model is now trajectory-independent. Meanwhile, the measurement model with reference frame attitude error definition can also be transformed into trajectory-independent form. Monte Carlo simulations were conducted to evaluate the MEKF variants with different state-space models. The results have shown that the performance of MEKF can be much improved with the trajectory-independent measurement models in some severe scenarios such as large initial estimation errors or erroneous initial error covariance.

## Acknowledgments

The authors would like to thank Prof. Haichao Gui from School of Astronautics, Beihang University for providing the numerical examples codes. The authors would like to thank Prof. Gongmin Yan from Northwestern Polytechnical University for providing the field test data.

## References

[1] Y. Wonkeun, "Magnetic Fault-Tolerant Navigation Filter," *IEEE Sensors Journal*, vol. 20, no. 22, pp. 13480-13490, 2020.

[2] W. X. Wang, and P. G. Adamczyk, "Comparison of Bingham Filter and Extended Kalman Filter in IMU Attitude Estimation," *IEEE Sensors Journal*, vol. 19, no. 18, pp. 8845-8854, 2019.

[3] M. B. D. Rosario, H. Khamis, P. Ngo, N. H. Lovell, and S. J. Redmond, "Computational Efficient Adaptive Error-State Kalman Filter for Attitude Estimation," *IEEE Sensors Journal*, vol. 18, no. 22, pp. 9332-9342, 2018.

[4] E. J. Lefferts, F. L. Markley, and M. D. Shuster, "Kalman Filtering for Spacecraft Attitude Estimation," *Journal of Guidance, Control, and Dynamics*, vol. 5, no. 5, pp. 417–429, 1982.

[5] F. L. Markley, "Attitude Error Representations for Kalman Filtering," *Journal of Guidance, Control, and Dynamics*, vol. 26, no. 2, pp. 311–317, 2003.

[6] J. L. Crassidis, F. L. Markley, and Y. Cheng, "Survey of Nonlinear Attitude Estimation Methods," *Journal of Guidance, Control, and Dynamics*, vol. 30, no. 1, pp. 12–28, 2007.

[7] Z. B. Qiu, and L. Guo, "Improved Cubature Kalman Filter for Spacecraft Attitude Estimation," *IEEE Transactions on Instrumentation and Measurement*, vol. 70, DOI: 10.1109/tim.2020.3041077, 2021.

[8] L. B. Chang, F. J. Qin, and F. Zha, "Pseudo Open-loop Unscented Quaternion Estimator for Attitude Estimation," *IEEE Sensors Journal*, vol. 16, no. 11, pp. 4460-4469, 2016.

[9] M. S. Andrle, and J. L. Crassidis, "Attitude Estimation Employing Common Frame Error Representations," *Journal of Guidance, Control, and Dynamics*, vol. 38, no. 9, pp. 1614–1624, 2015.

[10] A. Barrau, Non-linear State Error based Extended Kalman Filters with Applications to Navigation. Mines Paristech, 2015.

[11] A. Barrau, and S. Bonnabel, "Intrinsic Filtering on Lie Groups with Applications to Attitude Estimation," *IEEE Transactions on Automatic Control*, vol. 60, no. 2, pp. 436-449, 2014.

[12] A. Barrau, and S. Bonnabel, "The Invariant Extended Kalman Filter as a Stable Observer," *IEEE Transactions on Automatic Control*, vol. 62, no. 4, pp. 1797-1812, 2016.

[13] M. Brossard, A. Barrau, and S. Bonnabel, "Exploiting Symmetries to Design EKFs With Consistency Properties for Navigation and SLAM," *IEEE Sensors Journal*, vol. 19, no. 4, pp. 1572–1579, Sep. 2019.

[14] S. Heo and C. Park, "Consistent EKF-based Visual-Inertial Odometry on Matrix Lie Group," *IEEE Sensors Journal*, vol. 18, no. 9, pp. 3780–3788, May 2018.

[15] S. Heo, J. H. Jung, and C. G. Park, "Consistent EKF-based Visual Inertial Navigation Using Points and Lines," *IEEE Sensors Journal*, vol. 18, no. 18, pp. 7638–7649, Sep. 2018.

[16] H. Gui , A. H. J. de Ruiter, "Quaternion Invariant Extended Kalman Filtering for Spacecraft Attitude Estimation," *Journal of Guidance, Control, and Dynamics*, vol. 41, no. 4, pp. 863-878, 2018.

[17] L. B. Chang, , "SE(3) based Extended Kalman Filter for Spacecraft Attitude Estimation," *arXiv preprint arXiv:2003.12978*, 2020.

[18] R. Hartley, M. Ghaffari, R. M. Eustice, and J. W. Grizzle, "Contact-Aided Invariant Extended Kalman Filtering for Robot State Estimation," *The International Journal of Robotics Research*, 2020.

[19] J. L. Crassidis, and J. L. Junkins, Optimal Estimation of Dynamic Systems, 2nd ed., CRC Press, Boca Raton, FL, 2012, pp. 153–154, Chap. 3, 421-424, Chap. 7.

[20] L. B. Chang, F. J. Qin, and J. N. Xu, "Strapdown Inertial Navigation System Initial Alignment based on Group of Double Direct Spatial Isometries," *arXiv preprint arXiv:2102.12697*, 2021.

[21] M. Li, and A. I. Mourikis, "Improving the Accuracy of EKF-based Visual-inertial Odometry," *2012 IEEE International Conference on Robotics and Automation*. IEEE, 2012: 828-835.

[22] E. Gai, K. Daly, J. Harrison, and L. Lemos, "Star-Sensor-Based Satellite Attitude/Attitude Rate Estimator," *Journal of Guidance, Control, and Dynamics*, vol. 8, no. 5, pp. 560–565, 1985.

[23] M. W. Mueller , M. Hehn, and R. D'Andrea, "Covariance Correction Step for Kalman Filtering with an Attitude," *Journal of Guidance, Control, and Dynamics*, vol. 40, no. 9, Special Issue on the Kalman Filter and Its Aerospace Applications, pp. 2301–2306, 2017.

[24] L. B. Chang, J. S. Li and S. Y. Chen, "Initial Alignment by Attitude Estimation for Strapdown Inertial Navigation Systems," *IEEE Transactions on Instrumentation and Measurement*, vol. 64, no. 3, pp. 784-794, 2015.

[25] Y. X. Wu, H. L. Zhang, M. P. Wu, X. P. Hu, and D. P. Hu, "Observability of SINS alignment: A global perspective," *IEEE Transactions on Aero-space and Electronic Systems*, vol. 48, no. 1, pp. 78–102, 2012.